\documentclass[10pt,twocolumn,letterpaper]{article}

\usepackage[final,algorithms]{wacv}      

\usepackage{epsfig}
\usepackage{graphicx}
\usepackage{amsmath}
\usepackage{amssymb}
\usepackage{booktabs}
\usepackage{enumitem}

\usepackage{diagbox}
\usepackage{multirow}
\usepackage{soul}
\usepackage{color}
\usepackage{array}
\usepackage{ragged2e}
\usepackage{dblfloatfix}
\usepackage{float}

\usepackage{rotating}
\usepackage{makecell}
\usepackage{tabularray}
\usepackage{pifont}
\usepackage[bottom]{footmisc}

\usepackage[pagebackref,breaklinks,colorlinks,citecolor=magenta]{hyperref}

\usepackage[capitalize]{cleveref}
\crefname{section}{Sec.}{Secs.}
\Crefname{section}{Section}{Sections}
\Crefname{table}{Table}{Tables}
\crefname{table}{Tab.}{Tabs.}

\def\wacvPaperID{2124} 
\def\confName{WACV}
\def\confYear{2024}

\begin{document}

\title{PECoP: Parameter Efficient Continual Pretraining for \\ Action Quality Assessment}

\author{%
    Amirhossein Dadashzadeh\textsuperscript{1}, %
    Shuchao Duan\textsuperscript{1}, %
    Alan Whone\textsuperscript{2}, %
    Majid Mirmehdi\textsuperscript{1}\\
    \textsuperscript{1}School of Computer Science \hspace{0.2cm} \textsuperscript{2}Translational Health Sciences \\
    University of Bristol, UK\\
    {\tt\small \{a.dadashzadeh, shuchao.duan, alan.whone, m.mirmehdi\}@bristol.ac.uk}
}

\maketitle

\begin{abstract}

The limited availability of labelled data in Action Quality Assessment (AQA), has forced previous works to fine-tune their models pretrained on large-scale {domain-general} datasets. This common approach results in weak generalisation, particularly when there is a significant domain shift. We propose a novel, parameter efficient, continual pretraining  framework, PECoP, to reduce such domain shift via an additional pretraining stage. In PECoP, we introduce 3D-Adapters, inserted into the pretrained model, to learn spatiotemporal, in-domain information via self-supervised learning where only the adapter modules' parameters
are updated.  We demonstrate PECoP's ability to enhance the performance of recent state-of-the-art methods {(MUSDL, CoRe, and TSA)} applied to AQA, leading to considerable improvements on benchmark datasets, JIGSAWS ($\uparrow6.0\%$), MTL-AQA ($\uparrow0.99\%$), and FineDiving {($\uparrow2.54\%$)}. We also present a new Parkinson's Disease dataset, PD4T, of real patients performing four various actions, where we surpass ($\uparrow3.56\%$) the state-of-the-art in comparison. Our code, pretrained models, and the PD4T dataset are available at
{https://github.com/Plrbear/PECoP}.
\end{abstract}


\section{Introduction}
\label{sec:intro}


\begin{figure}[t]
\centerline{\includegraphics[scale=0.39]{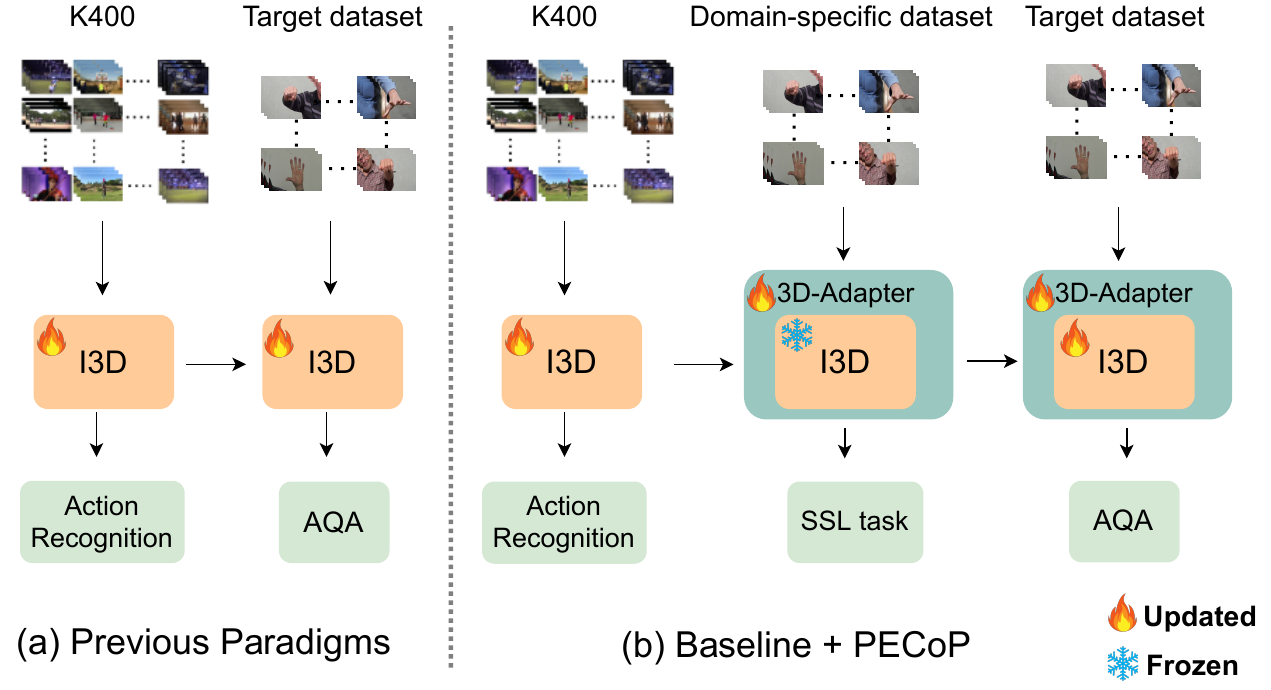}}
\caption{{(a) Previous works directly transfer the model pretrained on domain-general data to AQA downstream tasks with target data fine-tuning,  (b) in our proposed PECoP framework, the pretrained model continues to learn towards a specific AQA task through an additional pretraining stage, where only a small set of 3D-Adapter parameters are updated on unlabeled domain-specific data in a SSL approach, while the baseline model's weights remain frozen.} 
}
\label{Fig:fig-intro} 
\end{figure}

The analysis and evaluation of the quality of human action performance have a rich application in various real-world scenarios, such as sports events \cite{tang2020uncertainty,yu2021group, xu2022finediving}, healthcare \cite{liu2021towards,dadashzadeh2021exploring}, and skill assessment \cite{parmar2021piano, parmar2022domain}. 
Despite recent advances \cite{aqa-7, xu2019learning, tang2020uncertainty, yu2021group, lee2020biobert, dadashzadeh2021exploring, xu2022finediving, guo2022tree}, AQA methods are affected by insufficient quantities of annotated data for training deep networks \cite{aqa-7}. This is further exacerbated when extra effort is needed to produce very precise labels, e.g. for health-related applications, such as Parkinson's Disease (PD) severity assessment \cite{dadashzadeh2021exploring, lu2021quantifying, endo2022gaitforemer,morgan2022real}.
A common solution to address such problems is to start with a model that is originally pretrained on a large source dataset, commonly Kinetics-400 \cite{k-400} (K400), and finetune it on one's target AQA dataset \cite{tang2020uncertainty, yu2021group, xu2022finediving} (see Fig \ref{Fig:fig-intro}(a)). While better than training from scratch, deploying such pretrained models 
is sub-optimal for AQA,  due to the domain/task discrepancy between action classification and action assessment. For example, in PD action performance scoring,  one or two interruptions during the regular rhythm of a patient's movement performing an action can result in a different quality score and this is in contrast to the source pretraining task, where subtle or even more pronounced differences in performing an action should not affect action classification \cite{lei2019survey}. 


A promising route to address the shortcomings of this direct jump from classical pretraining to finetuning can be to further pretrain using domain-specific unlabeled data, i.e. continual pretraining --  
a strategy that has had a remarkable impact  in natural language processing (NLP) \cite{han2019unsupervised, gururangan2020don, wu2021domain} and recently in image/object classification \cite{reed2022self, azizi2022robust}.
When it comes to video-domain tasks (e.g. as in AQA), 
this additional pretraining stage on in-domain data may be computationally prohibitive or impractical, due to the requirement for updating all parameters and storing pretrained parameter sets for each separate task. 

BatchNorm tuning  \cite{frankletraining, reed2022self} could be used to
equip the pretrained model with domain-specific information by only updating the affine parameters of BatchNorm layers, while other pretrained parameters are frozen. Although this technique can greatly reduce the number of trainable parameters,
we show that in a continual pretraining framework it can fail on those AQA {datasets that are relatively small and more domain-specific, such as JIGSAWS \cite{gao2014jhu}.} 

In this paper, we propose adding a Parameter-Efficient Continual Pretraining (PECoP) adaptation stage to the traditional AQA transfer learning workflow that can efficiently {\it adapt} the domain-general pretrained model for the downstream AQA task. 
Inspired by adapter-based methods which have recently achieved strong results with transformer architectures on NLP benchmarks \cite{houlsby2019parameter, pfeiffer2020adapterhub, he2021towards, he2021effectiveness}, we present 3D-Adapter, a lightweight convolutional bottleneck block which is inserted into a pretrained 3D CNN (e.g. I3D inception modules \cite{carreira2017quo}) and learns domain-specific spatiotemporal knowledge via a self-supervised learning (SSL) approach. During domain-specific pretraining, only the adapter parameters are updated while the original weights of the pretrained model are frozen to allow a high degree of parameter-sharing (see Fig.~\ref{Fig:fig-intro}(b)). This greatly reduces the computational and storage costs of conventional continual pretraining, and also prevents overfitting by alleviating catastrophic forgetting \cite{he2021effectiveness}. 


To evaluate our method, we present experiments on three 
public AQA benchmarks,  
MTL-AQA \cite{parmar2019and}, JIGSAWS \cite{gao2014jhu} and FineDiving \cite{xu2022finediving}. We also provide comparative results on a new Parkinson's Disease dataset, PD4T, comprising four different PD motor-function tasks by real patients.  Our results clearly demonstrate that the proposed continual pretraining approach PECoP can boost the robustness of recent SOTA AQA methods, i.e. 
TSA \cite{xu2022finediving}, 
CoRe \cite{li2022pairwise} and {USDL/MUSDL} \cite{tang2020uncertainty}, 
by a considerable margin. 

In summary, the contributions of this work are: 
\begin{itemize}
    \item We propose a parameter-efficient continual pretraining workflow to enhance the efficiency of target tasks.
    \item We integrate a 3D-Adapter layer for the first time for 3D CNNs for video
analysis.
\item We introduce a new annotated AQA dataset,  PD4T, for the vision community to evaluate various actions performed by actual Parkinson's disease patients.
    \item  We present extensive experiments and ablations to demonstrate that recent state-of-the-art AQA methods can significantly benefit from our proposed continual pretraining approach.
\end{itemize}

\section{Related Work}
We address the most recent works that are significantly relevant to our work, including those on action quality assessment, self-supervised learning with respect to AQA tasks, continual pretraining, and adapters.

{\bf{AQA} --} Most AQA methods treat the task as a regression problem on various video representations supervised by the scores or labels given by expert judges 
\cite{parmar2019and, xu2019learning, liu2021towards, tang2020uncertainty, yu2021group, xu2022finediving}. 
To reduce the inherent ambiguity of such labels, Tang et al. \cite{tang2020uncertainty} proposed uncertainty-aware score distribution learning (USDL) for AQA. Yu et al. \cite{yu2021group} employed contrastive learning \cite{he2020momentum} and built a contrastive regression (CoRe) framework to learn relative scores by pair-wise comparison. Although these methods have achieved SOTA results on several AQA datasets, they disregard the significant domain gap between their base dataset (i.e. K400) and target AQA dataset.
In this paper, we address the pretraining stage towards better targeted learning in AQA methods.

{\bf{ SSL for AQA --}}
Although most of the SOTA works in the AQA literature have focused on supervised learning approaches, a few have recently explored self-supervised learning  \cite{roditakis2021towards, liu2021towards, li2022pairwise, zhang2022semi}. 
In these studies, in addition to the traditional supervised regression loss, the framework is further equipped with an SSL loss during the finetuning stage to improve the performance without the need for additional annotations. For instance, Liu et. al \cite{liu2021towards} employed a self-supervised contrastive loss to assist their supervised model capture temporal dynamics better in surgical videos. We leverage the advantages of SSL during the {\it{pretraining} process} to introduce to the pretrained model a level of domain-specific focus to allow it to handle the downstream AQA tasks more efficiently.

\begin{figure*}[t]

\centerline{\includegraphics[scale=0.59]{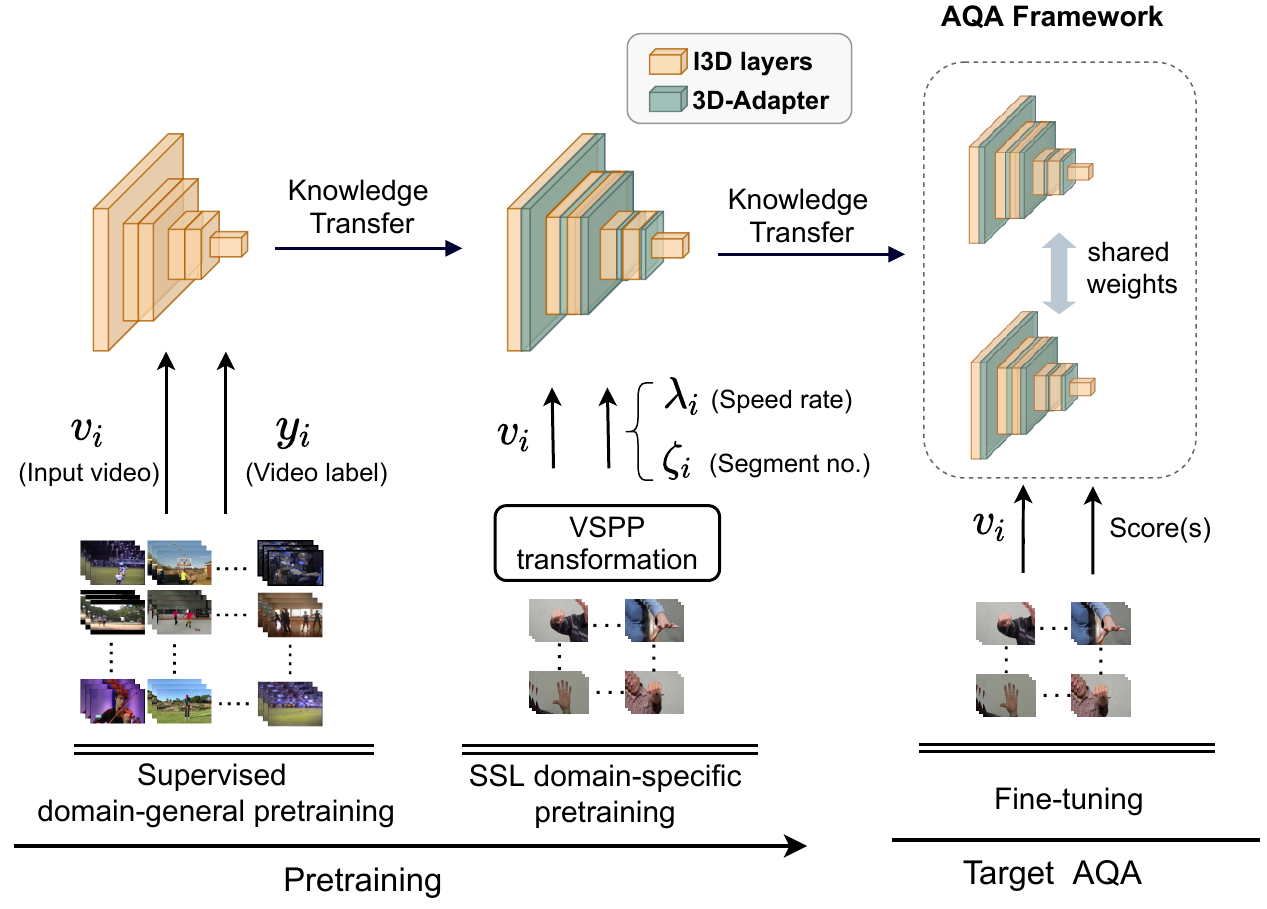}}
\caption{An overview of PECoP -- First, a 3D encoder is pretrained on a domain-general dataset (i.e. K400). Then, we equip the pretrained model with 3D-Adapters and update their parameters using VSPP \cite{dadashzadeh2022auxiliary}, a SSL pretext task, on unlabeled domain-specific data. Finally, we fine-tune the pretrained model on the AQA target task.}
\label{Fig:main}  
\end{figure*}

{\bf{{Continual pretraining --}}} 
In contrast to the traditional approach in transfer learning that follows domain-general pretraining (usually over ImageNet or K400), continual pretraining can enhance learning via in-domain self-supervised pretraining to handle domain shift problems 
\cite{gururangan2020don, pfeiffer2020adapterhub, wu2021domain, reed2022self, azizi2022robust, yang2022c3}.
Gururangan et .al \cite{gururangan2020don} showed the importance of an additional pretraining phase with in-domain data to improve their target task performance on text classification. In the image domain, 
Reed et. al \cite{reed2022self} verified that models continually pretrained on datasets that are progressively more similar to the target data can speed up convergence and increase robustness, while being particularly helpful when the target training data is limited.
Azizi et. al \cite{azizi2022robust} used a combination of both supervised pretraining on ImageNet, as well as intermediate contrastive SSL \cite{chen2020simple} on domain-specific medical data, to learn generalisable representations for medical images.
In this work, we investigate continual pretraining for the first time in the video domain, in particular on the AQA task. However, instead of continual pretraining of the entire  model parameter set, we take advantage of adapter-based tuning to reduce the cost of storage and model pretraining on in-domain data while preserving the knowledge obtained through the initial pretraining on domain-general data.


{\bf{{Adapters --}}}
Adapters are lightweight bottleneck modules which were designed for Transformer architectures for NLP tasks to conduct parameter-efficient transfer learning \cite{houlsby2019parameter, hu2021lora, he2021towards, ding2023parameter} for downstream tasks. In 
computer vision, Chen et al. \cite{chen2022adaptformer}
proposed AdaptFormer to efficiently adapt a pretrained vision transformer model \cite{dosovitskiy2020image} to scalable image and video recognition tasks. In \cite{chen2022conv}, Chen et al. introduced Conv-Adapter with a similar bottleneck architecture proposed for transformers, as in \cite{houlsby2019parameter, chen2022adaptformer}, 
but with convolution layers to enable adapters in 2D CNNs. To the best of our knowledge, there is no work yet to study the effect of Adapters in 3D CNNs, and our proposed approach fills this gap. 


\section{Proposed Approach}

Next, we outline our proposed continual pretraining approach implemented via self-supervised training of 3D-Adapter modules. The pipeline of our framework is shown in Fig. \ref{Fig:main}.




Let $D_g$ be a large-scale, annotated, domain-general video dataset used for a learning task $T_g$, and $D_t$ be a target video dataset in the AQA domain for a learning task $T_t$, with a significant domain discrepancy between $T_g$ and $T_t$. Then, given an unlabelled video dataset $D_q$, where $D_q \subseteq D_t$, our aim is to leverage the representations in $D_g$ and $D_q$ to learn a transferable spatiotemporal feature extractor that is able to perform as  well as possible on $D_t$ for task $T_t$.

{\bf{Domain-general pretraining --}} A pretrained backbone model, such as one that has been trained on a large video dataset, e.g. an I3D model supervised on K400, serves as our model architecture (see 1st column of Fig.~\ref{Fig:main}).

\begin{figure}[h]
\centerline{\includegraphics[scale=0.5]{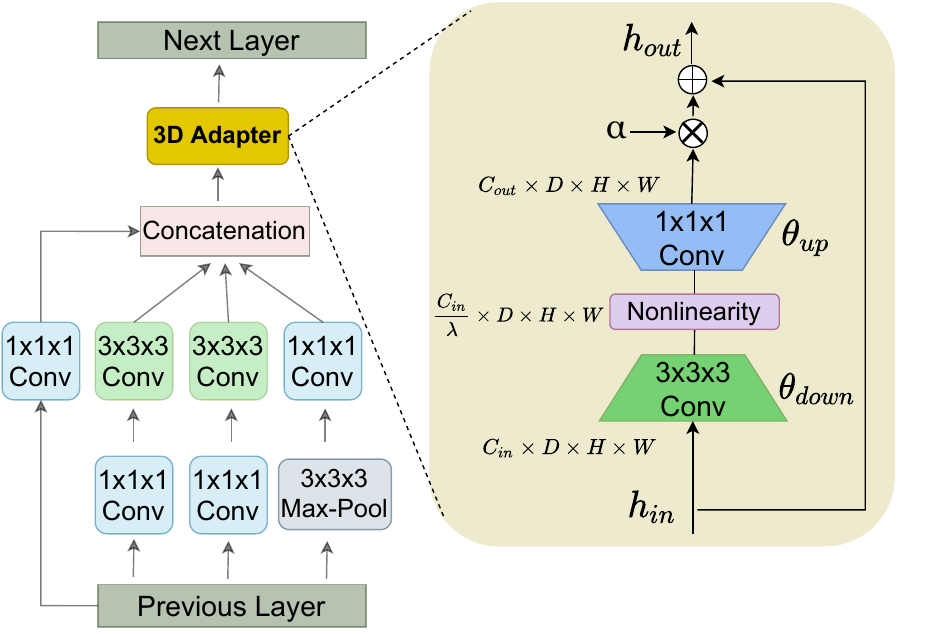}}
\caption{Inception module with adapter used in I3D model.}
\label{fig:3d-adpater}  
\end{figure}

{\bf{In-domain SSL continual pretraining -- }} We then equip our K400 pretrained model with randomly initialised 3D-Adapter modules. Our proposed 3D-Adapter has a similar bottleneck architecture as used in Transformers \cite{houlsby2019parameter, hu2021lora} and recently in 2D CNNs \cite{chen2022conv}, however, it requires 3D layers to be applied to 3D CNNs and trained on videos. The architecture of our 3D-Adapter and its integrated design with the inception module of the I3D model
is shown in Figure \ref{fig:3d-adpater}. A performance boost can be obtained if a single 3D-Adapter is inserted after the concatenation layer of each inception module.

A 3D-Adapter consists of a downsampling, depth-wise, 3D convolution with learnable weights $\theta_{down} \in \mathbb{R}^{\frac{C_{in}}{\lambda} \times\lambda \times K \times K \times K}$, a non-linear function $f(.)$, e.g. ReLU, followed by an upsampling, point-wise, 3D convolution with learnable weights  $\theta_{up} \in \mathbb{R}^{{C_{out}} \times \frac{C_{in}}{\lambda} \times 1 \times 1 \times 1}$. Here, $C_{in}$ and $C_{out}$ are the
channel dimensions of the input and output feature maps, respectively, $K=3$, and the compression factor $\lambda$ denotes the bottleneck's dimension. Hence, given an input feature vector $h_{in}$ $\in \mathbb{R}^{C_{in}\times D \times H \times W}$, then the output feature vector $h_{out}$ $\in \mathbb{R}^{C_{out}\times D \times H \times W}$ of our 3D-Adapter is
 \begin{equation}
     h_{out} = \alpha \odot(\theta_{up}\otimes f(\theta_{down}\bar{\otimes} h_{in})) + h_{in} ~,
 \end{equation}
 where  $\otimes$ and $\bar{\otimes}$ are point-wise and depth-wise 3D convolution respectively, and $\alpha$ is a tunable scalar hyperparameter in $\mathbb{R}^{C_{out}}$ which is initialised as ones, {and \( \odot \) denotes element-wise multiplication}, following \cite{hu2021lora, chen2022conv}. 
 

\begin{figure*}[!h]

\centerline{\includegraphics[scale=0.328]{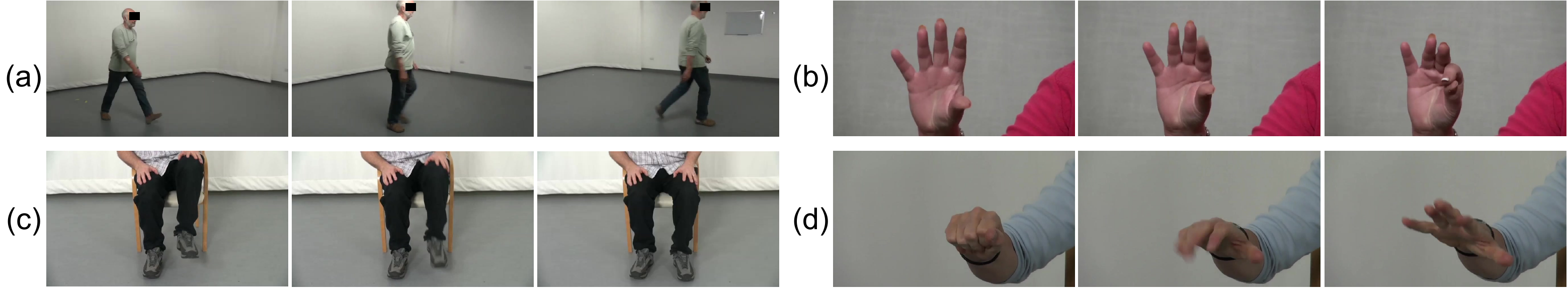}}
\caption{{Sample frames from the PD4T dataset: (a) gait, (b) finger tapping, (c) leg agility, and (d) hand movement. All videos are from actual PD patients and were captured in a  clinical environment as part of a clinical experiment across several months. For hand movement, finger tapping, and leg agility, data were collected from both the left and right sides for each subject.} 
}
\label{Fig:PD4T}  
\end{figure*}

During the proposed continual pretraining stage, we only allow the 3D-Adapter parameters to be optimised on $D_q$, while the original model layers' weights stay frozen (middle column in Fig. \ref{Fig:main}). The training is accomplished through self-supervised learning, with labels automatically generated from unlabeled videos in $D_q$. Since understanding the quality of action is heavily dependent on movement patterns, we focus on Video Segment Pace Prediction (VSPP) \cite{dadashzadeh2022auxiliary} as an SSL pretext task for this stage. VSSP requires our model to temporally explore a video clip and predict the index and speed of a segment within a clip that is sampled at a different speed rate. It was shown in \cite{dadashzadeh2022auxiliary} that VSPP is more suited to video motion events than previous video playback prediction tasks \cite{epstein2020oops, benaim2020speednet, wang2020self, chen2021rspnet} by way of much faster convergence. We shall explore the performance of other recent SSL methods (e.g. VideoPace \cite{wang2020self} and RSPNet \cite{chen2021rspnet}) during our ablations.


{\bf{Supervised fine-tuning --}} 
We select SOTA action quality assessment models
 CoRe \cite{yu2021group}, {USDL/MUSDL} \cite{tang2020uncertainty}, 
and TSA \cite{xu2022finediving},  
as example models that can be enhanced with  PECoP  and then fine-tuned on AQA datasets for direct evaluation. 
In essence, in this stage, each model's layers, including the original and the introduced adapter layers, are fine-tuned on the target dataset $D_t$ (see rightmost column in Fig. \ref{Fig:main}). 

For {\bf USDL}~\cite{tang2020uncertainty}, the features obtained from segments of a video clip pass through our continually pretrained I3D backbone and fused through temporal pooling,  and then sent through softmax to generate the predicted quality assessment distribution. The KL loss between the predicted distribution and a Gaussian distribution generated from the ground-truth score is applied for optimisation. \textbf{MUSDL} is a multi-path version of USDL which predicts the final score if multiple-judge scores are available, as is the case in the MTL-AQA and JIGSAWS datasets.

For {\bf CoRe}~\cite{yu2021group}, given pairwise query and exemplar videos, their spatiotemporal features  are extracted  with our continual pretraining I3D and then the two feature sets are combined with the reference score of the exemplar video and passed to the group-aware regression tree to obtain the score difference between the two videos. During inference, the final score is computed by averaging the results from multiple different exemplars.

For {\bf TSA} \cite{xu2022finediving}, spatiotemporal features are extracted similarly to CoRe. Then, a temporal segmentation attention module assesses action quality. 


\section{Experiments}
We evaluate our self-supervised continual pretraining adaptation module on the following datasets that span various AQA tasks:


(i) {\bf{MTL-AQA}} \cite{parmar2019and} {contains} 1412 video clips collected from 16 different {world} events and includes a variety of diving actions, covering both individual and synchronous divers, with videos from different angles. The annotations comprise scores from 7 judges, final scores, difficulty degree and type of diver’s action.


(ii) {\bf{JIGSAWS }}\cite{gao2014jhu} is a collection of 103 surgical activities that cover three distinct tasks: 39 Suturing (S), 28 Needle-Passing (NP), and 36 Knot-Typing (KT). Each task is annotated by multiple sub-scores (that represent e.g., flow of operation, quality of final outcome, and so on), with the final score defined as the sum of these sub-scores.
%

(iii) {\bf{FineDiving }}\cite{xu2022finediving} is a
fine-grained sports video dataset for AQA which provides 3,000 video of diving with detailed annotations on 52 actions, 29 sub-actions, and 23 difficulty degree types.

(iv) {\bf{PD4T}} is our new fully annotated dataset offering 2931 videos from 30 PD patients tested longitudinally at 8 week intervals. {The videos were captured at 25fps at a resolution of 1920×1080
(reduced to 854x480), using a SONY HXR-NX3 camera. There are a total of 30 subjects, with 22 of these used for training and 8 for testing.}
The patients, who ranged from 41 to 72 years old,  performed PD tasks of gait, finger tapping, hand movement, and leg agility in clinical settings and their Unified Parkinson's Disease Rating Scale (UPDRS) \cite{goetz2008movement} quality scores were assigned by trained clinicians ranging from 0 (normal) to 4 (severe)\footnote{For additional information about the PD4T dataset, see the appendix.}. Sample frames are shown in Figure~\ref{Fig:PD4T}.

~\\ \noindent {\bf Experiment Setup --}
The experiments were performed on an
Nvidia RTX 3090TI GPU under Cuda 11.6 with cuDNN
8.2. We first initialise our I3D model with K400 pretrained weights which then remain frozen throughout the pretraining stage. After adding 3D-Adapters, the classification head is replaced with two randomly initialised FC layers $f_{\lambda}$ and $f_{\zeta}$ corresponding to the segment speed and index outcomes of the VSPP~\cite{dadashzadeh2022auxiliary} pretext task. Different values for $f_{\lambda}$ and $f_{\zeta}$ are used for different AQA tasks. We perform SSL pretraining on domain-specific datasets by only updating the 3D-Adapter layers over 8 epochs, with batch size of 16 and SGD with a $1 \times 10^{-3}$ learning rate. {In this stage, we generate 32-frame long video clips, and  empirically set the two parameters needed for VSSP to $[\lambda=4, \zeta=4]$ or $[\lambda=4, \zeta=3]$ which is either at, or close to, those recommended in \cite{dadashzadeh2022auxiliary}.}{ Note that the training set videos of the target data is our domain-specific dataset for SSL pretraining.} 
For data augmentation, we randomly crop the video clips to $224\times224$ followed by horizontal flip and colour jittering of each frame. Following \cite{dadashzadeh2022auxiliary}, we apply 10x more iterations per epoch for temporal jittering.
In all experiments, the input clip length is 32 during pretraining. 

~\\ \noindent {\bf{Fine-tuning --}} 
The pretrained I3D model is then the backbone network of our baselines {USDL, MUSDL}, CoRe, and TSA, 
and we evaluate their performance with the {Spearman Rank Correlation} metric ($\mathcal{S}$), expressed as percentages. For the JIGSAWS dataset, in keeping with other methods \cite{tang2020uncertainty,yu2021group}, we provide values after four-fold cross-validation. In this stage, we adopt similar hyperparameter settings and training/evaluation strategy for each baseline as reported in  \cite {tang2020uncertainty}, \cite{yu2021group}, and \cite{xu2022finediving} 
respectively.

~\\ \noindent {\bf Comparative Evaluation --}
We present results on the MTL-AQA \cite{parmar2019and} and JIGSAWS \cite{gao2014jhu} datasets against SOTA AQA methods MUSDL \cite{tang2020uncertainty} and CoRe \cite{yu2021group} when we enhance them with PECoP, as well as when we enhance them with another recent continual pretraining workflow, HPT \cite{reed2022self} (see Table \ref{table:MTL-JIG}). In HPT,  which has only been applied to image domain tasks till now,  simply additional pretraining steps are introduced on domain-specific datasets with all model parameters updated at every stage. We use the same hyperparameters for training both PECoP and HPT.

While with PECoP improved results are obtained across the board, the improvements on JIGSAWS are very significant,
{e.g. after adding PECoP, MUSDL's average performance on the three tasks in the JIGSAWS dataset improves to 76\% ($\uparrow6\%$). Similarly, CoRe's average performance on the same tasks increases to 89\% ($\uparrow4\%$)}. This clearly shows PECoP's effectiveness in narrowing the substantial domain gap between the JIGSAWS dataset and K400.

Further, we note that adding HPT to the baselines results in a performance drop on JIGSAWS. Specifically, CoRe's performance decreases to 80\% ($\downarrow5\%$). This decline can be attributed to overfitting, as HPT requires all model parameters to be pretrained on a relatively small dataset (i.e. $\sim$13M parameters vs. PECoP's 3D-Adapters' $\sim$1M).

\begin{table*}[h]
\centering

\caption{Spearman Rank Correlation results on MTL-AQA and JIGSAWS, with and without continual pretraining. $~^\star$ ViSA \cite{li2022surgical} and MultiPath-VTPE \cite{liu2021towards} are customised towards surgical skill assessment and not general AQA tasks.} 
\label{table:MTL-JIG}
\begin{tabular}{l|l|c||ccc|c} 
\hline
\multirow{2}{*}{\textbf{Method}} & \multirow{2}{*}{ \textbf{Year}} & \textbf{MTL-AQA} & \multicolumn{4}{c}{\textbf{JIGSAWS}} \\ 
\cline{3-7}
& & {\bf Diving} & {\bf S} & {\bf NP} & {\bf KT} & {\bf Avg $\mathcal{S}$} \\

\hline

{USDL \cite{tang2020uncertainty}} & 2020& 90.66 & 64 & 63 & 61 & 63 \\
{MultiPath-VTPE \cite{liu2021towards}}$^\star$ & 2021 & - & 82 & 76 & 83 & 80 \\
{TSA-Net \cite{wang2021tsa} } & 2021 & 94.22 & - & - & - & - \\
{I3D + MLP \cite{yu2021group}} & 2021& 89.21 & 61 & 68 & 66 & 65 \\
{I3D-TA \cite{zhang2022learning} } & 2022 & 92.79 & - & - & - & - \\
{ViSA\cite{li2022surgical}}$^\star$ &2022 & - & 84 & 86 & 79 & 83 \\
{ResNet34-(2+1)D-WD \cite{farabi2022improving}} &2022 & 93.15 & - & - & - & - \\

\hline
{MUSDL \cite{tang2020uncertainty}} &2020 & 92.73 & 71 & 69 & 71 & 70 \\
{MUSDL + HPT \cite{reed2022self}} & 2023 & 93.49 & 69 & 75 & 72 & 72 \\
{MUSDL + PECoP} &2023 & \bf{93.72} & {\bf 77} & {\bf 76} & {\bf 76} & \bf{76} \\
\hline

{CoRe \cite{yu2021group} } &2021 & 95.12 & 84 & 86 & 86 & 85 \\
{CoRe + HPT \cite{reed2022self}} & 2023& 94.26 & 80 & 81 & 80 & 80 \\
{CoRe + PECoP} &2023 & \bf{95.20} & {\bf 88} & {\bf 90} & {\bf 88} & \bf{89} \\ 
\hline
\end{tabular}
\end{table*}

\begin{table}[h]
\centering

\caption{Comparison of PECoP and HPT \cite{reed2022self} in terms of storage size and pretraining cost.}
\label{table:Params}
\begin{tabular}{l|c|c|c} 
\hline
\begin{tabular}[c]{@{}l@{}} {\bf Continual}\\ {\bf Pretraining}  \end{tabular}        & \begin{tabular}[c]{@{}l@{}}{\bf \#trainble}\\{\bf parameters}\end{tabular} & {\bf \#epochs}  & {\bf Size}    \\ 
\hline
HPT \cite{reed2022self}       & $\sim$13M                                                            & 16                   & $\sim$54MB                  \\
PECoP         & $\sim$1M                                                             & 8                    & $\sim$4MB                    \\ 
\hline
  
\end{tabular}
\end{table}

\begin{table}[h]
\centering

\caption{Results on FineDiving Dataset.}
\label{Table:FD}
\begin{tabular}{l|cc} 
\hline
{\bf Method}     & $\mathcal{S}$    &          \\ 
\hline

CoRe \cite{yu2021group}     & 90.61         &      \\
CoRe + PECoP     & \bf{93.15}        &        \\
TSA \cite{xu2022finediving}     & 92.03     &      \\
TSA + PECoP & \multicolumn{1}{c}{{{93.13}}} & \multicolumn{1}{c}{}  \\
 \hline
\end{tabular}
\end{table}

\begin{table*}[t]

\centering
\renewcommand\arraystretch{1.2}
\caption{Spearman Rank Correlation results on the  PD4T dataset.}
\label{Table:PD4T}
\begin{tabular}{l|cccc|c} 
\hline
 \textbf{Method} & {\bf Gait} & {\bf Finger tapping} & {\bf Hand movem.} & {\bf Leg agility} & {\bf Avg. $\mathcal{S}$} \\ 
\hline
{USDL \cite{tang2020uncertainty}} & 79.14 & 42.58 & 53.93 & 56.47 & 58.03 \\
{USDL + HPT \cite{reed2022self}} & {\bf 81.93} & 46.38 & 54.15 & {\bf 58.54} & \bf{60.25} \\
{USDL + PECoP} & 80.68 & {\bf 47.44} & {\bf 56.19} & 58.09 & 60.06 \\
\hline
{CoRe \cite{yu2021group}} & 78.87 & 45.93 & 54.10 & 62.34 & 60.31 \\
{CoRe + HPT \cite{reed2022self}} & 81.42 & {\bf 49.73} & 57.06 & 63.98 & 63.05 \\
{CoRe + PECoP} & {\bf 82.33} & 49.40 & {\bf 59.46} & {\bf 64.27} & \bf{63.87} \\ 
\hline
\end{tabular}
\end{table*}

{In Table \ref{table:Params}, we show that not only PECoP dramatically reduces the {trainable parameters}, it also requires drastically 
fewer epochs to converge compared with HPT.      
}



{Table \ref{Table:FD} presents the results on the FineDiving dataset introduced in \cite{xu2022finediving},  comparing CoRe \cite{yu2021group} and TSA \cite{xu2022finediving}, with and without PECoP. Since TSA requires step transition labels for training, we cannot evaluate it on other datasets. 
As shown, TSA+PECoP improves on TSA by $1.10\%$. Although TSA outperforms CoRe alone, CoRe+PECoP surpasses TSA and TSA+PECoP to achieve the SOTA performance on FineDiving dataset. }


Table \ref{Table:PD4T} presents the results for the PD4T dataset. Given only a single action performance score based on the UPDRS scale~\cite{goetz2008movement} is available per clip, we compare PECoP and HPT for USDL instead of MUSDL. 
We observe  the Spearman’s rank correlation improves  when averaged across the four PD4T tasks for both HPT and PECoP when added to both USDL and CoRe ($\uparrow2.03\%$ and $\uparrow3.56\%$ respectively for PECoP ), although HPT performs marginally better ($\uparrow2.22\%$) when added to USDL.
{We assume this slight advantage for HPT is likely due to its greater model capacity for handling the complex patterns in the PD4T dataset; however, PECoP achieves nearly equivalent performance gains while significantly reducing continual pretraining and storage costs.}







\begin{figure*}[!b]
\centerline{\includegraphics[scale=0.43]{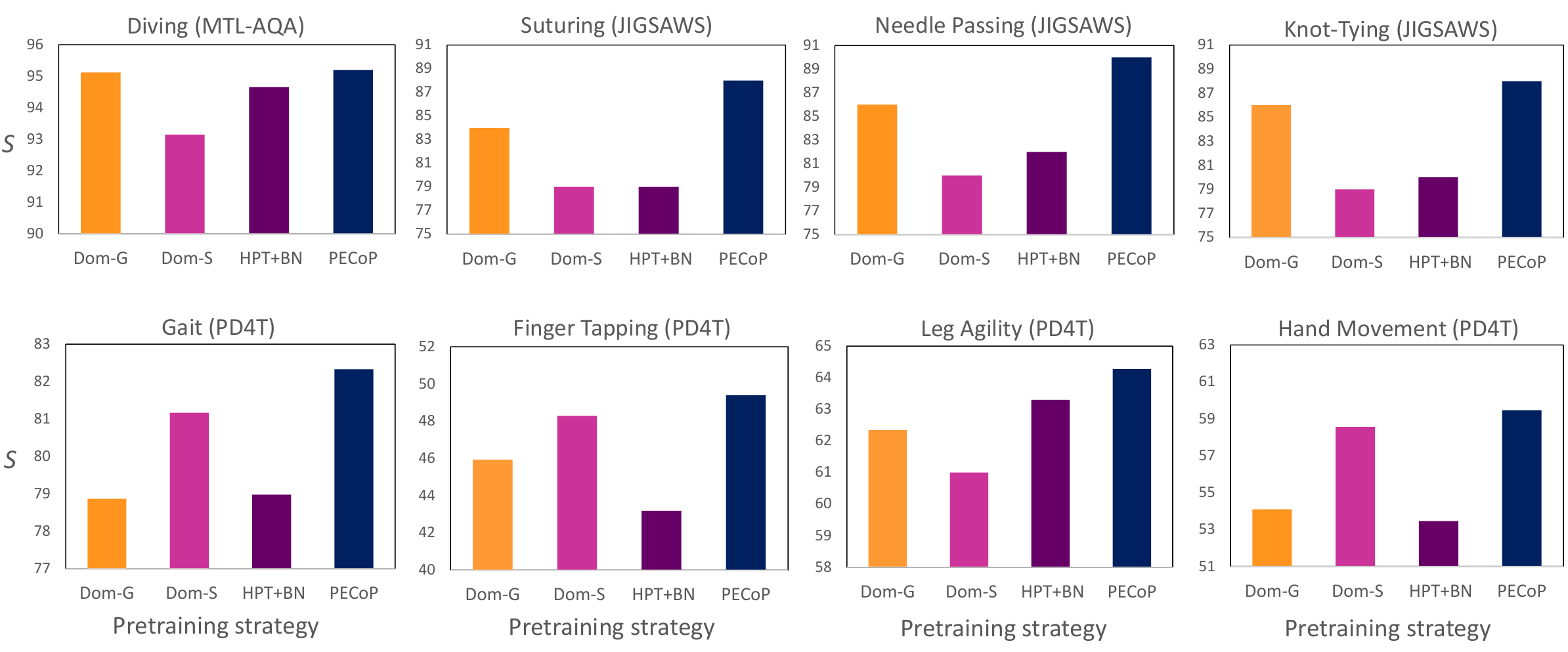}}
\caption{{Comparison of PECoP with Domain-Specific SSL Pretraining (Dom-S), Domain-General Pretraining (Dom-G), and BatchNorm Tuning (HPT+BN) across eight different AQA tasks from MTL-AQA, PD4T, and JIGSAWS datasets. Each plot represents a unique AQA task and shows the performance of the four approaches. Dom-G employs pretraining on a domain-general dataset like K400, while Dom-S focuses on domain-specific self-supervised pretraining on target data. HPT+BN fine-tunes only the BatchNorm layers of a pretrained model. PECoP consistently outperforms the other approaches across all tasks, indicating its robustness and adaptability for AQA tasks with different domains and complexities. }}


\label{Fig:BatchNorm}  
\end{figure*}






\begin{table}[!t]

\centering
\caption{Determining which SSL pretext task would be better to use - comparing contrastive learning approach to transformation-based ones. The experiment was performed on  the JIGSAWS dataset as an example.}
\label{Table:SSLs}
\begin{tabular}{l|ccc|c} 
\hline
{\bf Method} & {\bf S} & {\bf NP} & {\bf KT}     & {\bf Avg.} $\mathcal{S}$                                    \\ 
\hline
RSPNet \cite{chen2021rspnet}   &     83 &  86 &    84    & 84         \\
VideoPace\cite{wang2020self}    &     86 & 87  & 87   &  87           \\
VSPP \cite{dadashzadeh2022auxiliary}     &  \bf{88}    & \bf{90} &    \bf{88}   & \bf{89}              \\
\hline
\end{tabular}
\end{table}

\section{Ablation Study and Analysis}
{In this section, we perform ablations on our PECoP framework for AQA tasks, focusing on the influence of different SSL methods employed for domain-specific pretraining, the role of BatchNorm tuning in domain adaptation, and the impact of integrating 3D-Adapters into another 3D CNN, e.g. R3D-18\cite{hara2017learning}.}

\noindent {\bf{Different SSL Methods --}}
We investigate the performance of PECoP when using different SSL methods for domain-specific pretraining, i.e. RSPNet \cite{chen2021rspnet}, a contrastive learning-based SSL approach (based on MoCo \cite{he2020momentum}), and VideoPace \cite{wang2020self}, a transformation-based SSL pretex task (similar to VSPP). In this ablation, CoRe has been used as the AQA baseline and JIGSAWS as the target AQA task. As shown in Table \ref{Table:SSLs}, VSSP achieves the best result for domain-specific pretraining. 

{\bf{BatchNorm (BN) Tuning --}}
As mentioned earlier, BN tuning can be used to
equip a pretrained model with domain-specific knowledge by only updating the affine parameters of BatchNorm layers. Figure \ref{Fig:BatchNorm} illustrates the comparative performance of BatchNorm tuning (HPT+BN) and other pretraining strategies, such as domain-general pretraining (Dom-G), domain-specific SSL pretraining (Dom-S) from scratch, and PECoP. Again, CoRe is used as the AQA baseline.

Each plot corresponds to an AQA target task taken from our various datasets. On all these tasks, PECoP outperforms HPT+BN, particularly by a large margin on the tasks within {PD4T} and {JIGSAWS} datasets. Further, we observe that, HPT+BN performs significantly worse than Dom-G alone {on the all three tasks within JIGSAWS} dataset. 
This suggests that the BN affine parameters, $\beta$ and $\gamma$,
generally have a negative impact on downstream AQA tasks when facing a significant domain shift.
 {This happens because $\beta$ and $\gamma$ are tuned to the feature distribution of the source domain. When these parameters are applied to a significantly different target domain, they fail to correctly adjust feature statistics, resulting in a misalignment between the source and target domains. This issue becomes worse in smaller target datasets, as the limited number of examples makes it harder for the model to learn the true feature distribution, increasing the risk of overfitting.}

{In addition, the figure also shows that Dom-S performs variably. It fared worse than Dom-G on MTL-AQA, but exceeds Dom-G in nearly all PD4T tasks.
This discrepancy suggests that direct transfer from the K400 dataset may not be ideal for PD4T tasks due to a significant domain gap. On the other hand, Dom-S also performs poorly on all JIGSAWS tasks, implying that self-supervised pretraining from scratch on smaller datasets is sub-optimal. These observations highlight the need for adaptable pretraining strategies, an aim achieved by our PECoP approach, which consistently delivers superior performance in each of the evaluated tasks.}


\begin{table*}[h]

\centering
\renewcommand\arraystretch{1.2}
\caption{Spearman Rank Correlation results on the  PD4T dataset with R3D-18 backbone used in CoRe.}
\label{Table:r3d}
\begin{tabular}{l|cccc|c} 
\hline
 \textbf{Method} & {\bf Gait} & {\bf Finger tapping} & {\bf Hand movem.} & {\bf Leg agility} & {\bf Avg. $\mathcal{S}$} \\ 
\hline
{CoRe}  & 76.16 & 35.35 & {50.53} & 49.96 & 53.0 \\ 
{CoRe + PECoP} & {\bf 79.11} & {\bf{39.71}} & {\bf{55.37}} & {\bf 52.56} & {\bf 56.69} \\

\hline
\end{tabular}
\end{table*}

\begin{figure}[h]
\centerline{\includegraphics[scale=0.47]{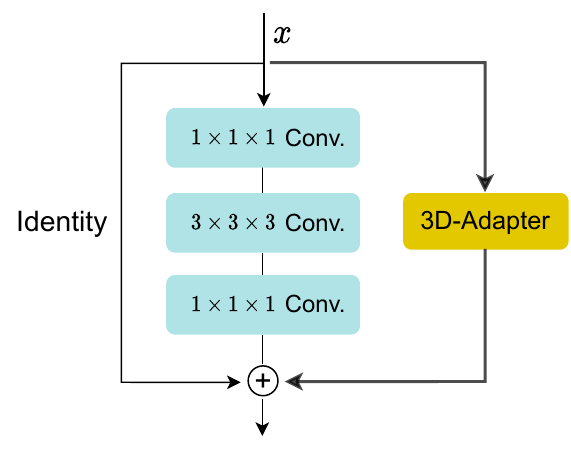}}
\caption{3D residual block equipped with 3D-Adapter used in the R3D-18 model. We empirically find that such a configuration leads to a better performance. }
\label{fig:3d-adpater-resnet}  
\end{figure}

{\bf{3D-Adapters for ResNet --}}
We evaluate the effectiveness of our 3D-Adapter with another 3D-CNN, i.e. R3D-18 \cite{hara2017learning} which is a common backbone network for action recognition tasks. We first insert a 3D-Adapter into each 3D residual blocks of R3D-18 backbone (see Fig. \ref{fig:3d-adpater-resnet}) and train this model through our continual pretraining framework, {PECoP}. This model is then used as the backbone network for CoRe to finetune on the target AQA task. We conduct this experiment on PD4T dataset and the results are reported in Table \ref{Table:r3d}. As shown, {across all four tasks within PD4T dataset}, CoRe+PECoP outperforms CoRe alone (i.e. CoRe with R3D-18 backbone in both cases).

\section{Discussion}
{\bf{Learning efficiency of PECoP --}}
{PECoP allows the model to leverage knowledge gained from a previous stage and requires only a small subset of parameters to be specifically learned for the new task. This approach leads to fewer epochs and less training data required for convergence on a new task, resulting in a reduction of both computational resources and pretraining time (refer to Table \ref{table:Params}). With fewer trainable parameters, PECoP has a limited capacity to memorise the training data, and as a result, it is forced to learn more generalised patterns, making it less prone to overfitting.}

{Another benefit to PECoP's efficiency is its ability to avoid the issue of forgetting \cite{he2021effectiveness}. Unlike traditional continual pretraining methods \cite{reed2022self,azizi2022robust} that require updating all model parameters, thereby risking forgetting previously learned patterns, PECoP leverages pre-existing domain-general knowledge without modifying this foundational knowledge. Instead, it simply augments it with domain-specific knowledge, thus effectively alleviating the issue of forgetting.
}

{\bf{Importance of PECoP for AQA --}}
{PECoP is crucial for AQA tasks, offering unique benefits that are valuable in diverse applications, ranging from healthcare settings like PD severity assessment to sports evaluations such as diving.} Its scalable design eliminates the need for multiple pretrained models for different assessments, a key advantage where quick and precise evaluations across a range of tasks are essential. Furthermore, PECoP's data efficiency makes it well-suited for healthcare settings where annotated data is often scarce. Its computational efficiency also facilitates faster decisions and saves resources, which is vital in environments with limited computational capabilities.

{\bf{Limitations and Future Work --}}
{One of the key weaknesses in a parameter-efficient continual pretraining ap-
proach like PECoP is the potential trade-off between efficiency and complexity in the model. Although using a limited set of adaptable parameters minimises the risk of overfitting, it may unintentionally lead to underfitting, making the model less effective at capturing intricate spatial and temporal features. This limitation becomes especially noticeable in AQA tasks, where accurately capturing subtle details may require a model with greater capacity.
Additionally, PECoP has primarily been evaluated using 3D CNN backbones for AQA tasks. The effectiveness of PECoP with other architectures, e.g. transformers, as well as its applicability to other vision tasks (e.g., action recognition and few shot learning), requires further validation. We plan to address these in our future work.
}

\section{Conclusions}
We proposed PECoP, a parameter efficient continual pretraining workflow  to better transfer the knowledge learned from existing large-scale video datasets (e.g. K400) to AQA target tasks by only updating a small number of additional bottleneck layers (called 3D-Adapters) through self-supervised learning. Alongside the evaluation on benchmark datasets, we also presented results on a new dataset of functional mobility actions performed by actual Parkinson's patients for performance quality assessment, with potential for longitudinal evaluation.  Experiments on four AQA datasets (8 different tasks) with three AQA baselines (CoRe, USDL/MUSDL, and TSA) demonstrated the significant advantages of PECoP over the conventional continual pretraining approach with respect to both generalisation ability, storage needs, and training cost.

\section{Acknowledgements}
{The authors would like to gratefully acknowledge the contribution of the Parkinson’s study participants. The 1st author is funded by a kind donation made to Southmead Hospital Charity, Bristol and by a donation from Caroline Belcher via the University of Bristol's Development and Alumni Relations. The clinical trial from which the video data of the people with Parkinson’s was sourced was funded by Parkinson’s UK (Grant J-1102), with support from Cure Parkinson’s.}

{\small
\bibliographystyle{ieee_fullname}
\bibliography{egbib}

}

\clearpage

\appendix
\section*{\Large{Appendix}}

\section{More details on PD4T}
The number of videos (\#video) for each score of this dataset, as well as the minimum/maximum number of frames (\#min/\#max) for each task can be seen in
Table \ref{Table:PD4Tdetails}.

For {\bf{gait}} analysis, the patients were asked to walk 10 metres at a comfortable pace and then return to their starting point. For  {\bf{hand movement}}, the patients opened and closed each of their hands (separately) 10 times, as fully and as quickly as possible. For {\bf{finger tapping}} each patient had to tap their index finger on their thumb 10 times quickly while spanning the amplest range possible. For {\bf{leg agility}}, while seated, the patient was asked to raise their foot high and stomp on the ground repeatedly and quickly for 10 times.

\begin{table}[!h]
\scriptsize
\centering
\caption{The PD4T dataset summary, categorized by severity scores. For each of
the four motor tasks the
table lists the total number of videos (\#video), the minimum (\#min) and maximum
(\#max) number of frames for the respective task.}
\label{Table:PD4Tdetails}
\begin{tabular}{c|l|rrrrr} 
\hline
\multicolumn{2}{c|}{Score} & \begin{tabular}[c]{@{}c@{}}Normal\\~(0)\end{tabular} & \begin{tabular}[c]{@{}c@{}}Slight \\(1)\end{tabular} & \begin{tabular}[c]{@{}c@{}}Mild \\(2)\end{tabular} & \begin{tabular}[c]{@{}c@{}}Moderate \\(3)\end{tabular} & \begin{tabular}[c]{@{}c@{}}Severe \\(4)\end{tabular} \\ 
\hline
\multirow{3}{*}{Gait} & \#video & 196 & 158 & 64 & 8 & 0 \\
 & \#min & 325 & 580 & 421 & 664 & - \\
 & \#max & 980 & 1866 & 13428 & 10688 & - \\ 
\hline
\multirow{3}{*}{\begin{tabular}[c]{@{}c@{}}Finger \\tapping\end{tabular}} & \#video & 152 & 465 & 164 & 23 & 2 \\
 & \#min & 129 & 129 & 129 & 162 & 159 \\
 & \#max & 450 & 724 & 853 & 398 & 460 \\ 
\hline
\multirow{3}{*}{\begin{tabular}[c]{@{}c@{}}Hand \\movem.\end{tabular}} & \#video & 234 & 407 & 179 & 23 & 5 \\
 & \#min & 131 & 136 & 150 & 197 & 220 \\
 & \#max & 334 & 571 & 717 & 648 & 648 \\ 
\hline
\multirow{3}{*}{\begin{tabular}[c]{@{}c@{}}Leg \\agility\end{tabular}} & \#video & 407 & 376 & 54 & 11 & 3 \\
 & \#min & 129 & 135 & 155 & 273 & 345 \\
 & \#max & 513 & 427 & 686 & 504 & 435 \\
\hline
\end{tabular}
\end{table}

\section{Temporal Parsing Transformer \cite{bai2022action}}



\begin{table}[b]
\scriptsize
\centering
\caption{Spearman Rank Correlation results on the  PD4T dataset with TPT as the baseline.}
\label{Table:TPT}
\begin{tabular}{l|cccc|c} 
\hline
\textbf{Method} & Gait & \begin{tabular}[c]{@{}c@{}}Finger \\tapping\end{tabular} & \begin{tabular}[c]{@{}c@{}}Hand \\movem.\end{tabular} & \begin{tabular}[c]{@{}c@{}}Leg \\agility\end{tabular} & \begin{tabular}[c]{@{}c@{}}Avg. \\$\mathcal{S}$\end{tabular} \\ 
\hline
TPT & 77.80 & 36.05 & 47.80 & 46.27 & 51.98 \\
TPT + PECoP & 79.90 & 40.73 & 51.07 & 50.38 & \bf{55.52} \\
\hline
\end{tabular}
\end{table}

The Temporal Parsing Transformer (TPT) is a recent SOTA AQA method based on transformers \cite{bai2022action}. 
Unlike existing AQA methods that focus on holistic video representations for score regression, TPT 
decomposes the video into temporal segments (part-level representations) to extract features. Such a decomposition is critical to TPT's  learning process to capture the possible phases of a typical AQA action, e.g. a diving action which contains several key parts, such as approach, take off, flight, etc.

We evaluate the performance of TPT\footnote{To train and evaluate TPT we used the code provided in https://github.com/baiyang4/aqa\_tpt } on our PD4T dataset with and without PECoP.
To this end, we first split the input video into 5 overlapping clips, and  feed each clip into our
continually pretrained I3D backbone to get clip level feature representations. Then, TPT is used to convert these representations into temporal part-level representations. Finally, a part-aware contrastive regressor (following \cite{yu2021group}) computes part-wise relative representations and fuses them to perform the final relative score regression.   


As shown in Table \ref{Table:TPT}, PECoP significantly boosts the performance of TPT across the various actions in PD4T. We note that, the performance of TPT is significantly lower than CoRe and USDL on PD4T tasks (See Table 4 in the main paper). We believe this may be attributed to the substantial degree of action repetition (e.g. in finger tapping or leg agility). In such cases, TPT's part-level representations, as opposed to a more holistic representation, cannot provide enough discriminative information for its learning process and hence TPT's part-level representations do not necessarily align well with some AQA tasks, such as those in PD4T. 





\end{document}